\documentclass[10pt,twocolumn,letterpaper]{article}
\usepackage[final]{cvpr} 
\usepackage{graphicx}
\usepackage{mathtools} 
\usepackage{graphicx}
\usepackage{amsmath}
\usepackage{amssymb}
\usepackage{booktabs}
\usepackage{dsfont}
\usepackage{multirow}
\usepackage{multicol}
\usepackage[pagebackref,breaklinks,colorlinks]{hyperref}
\usepackage[table]{xcolor}
\usepackage[accsupp]{axessibility}
\usepackage{adjustbox}
\usepackage[capitalize]{cleveref}
\crefname{section}{Sec.}{Secs.}
\Crefname{section}{Section}{Sections}
\Crefname{table}{Table}{Tables}
\crefname{table}{Tab.}{Tabs.}

\usepackage{setspace}

\usepackage{amsmath,amsfonts,bm}
\usepackage{pifont}
\usepackage{dsfont}

\newif\ifshowedits

\newcommand{\addeditor}[3]{%
  \definecolor{#1color}{rgb}{#3}
  \expandafter\newcommand\csname #1\endcsname[1]{%
  \ifshowedits
    {\color{#1color} ##1}%
  \else
    {##1}%
  \fi
  }%
  \expandafter\newcommand\csname #1rmk\endcsname[1]{%
  \ifshowedits
    {\color{#1color} {\bf [#2: ##1]}}
  \fi
  }%
  \expandafter\newcommand\csname #1rpl\endcsname[2]{%
  \ifshowedits
    {{\color{#1color} ##1} \sout{##2}}
  \else
    {##1}
  \fi
  }%
}

\def\eqref#1{equation~\ref{#1}}

\def\1{\bm{1}}

\def\rve{{\mathbf{e}}}

\DeclareMathAlphabet{\mathsfit}{\encodingdefault}{\sfdefault}{m}{sl}
\SetMathAlphabet{\mathsfit}{bold}{\encodingdefault}{\sfdefault}{bx}{n}

\newcommand{\relativeR}[0]{\Delta R}

\newcommand{\MedErr}[0]{\text{Median}}

\newcommand{\AccThirty}[0]{\text{Acc 30}}

\def\cvprPaperID{4790} 
\def\confName{CVPR}
\def\confYear{2024}
\title{NOPE: Novel Object Pose Estimation from a Single Image
} 
\author{Van Nguyen Nguyen$^{1}$,  Thibault Groueix$^{2}$, Georgy Ponimatkin$^{1}$, Yinlin Hu$^{3}$, Renaud Marlet$^{1,4}$,\\
 Mathieu Salzmann$^{5}$,   Vincent Lepetit$^{1}$\\ 
\\
{$^{1}$LIGM, Ecole des Ponts}, {$^{2}$Adobe}, {$^{3}$MagicLeap}, {$^{4}$Valeo.ai}, {$^{5}$EPFL}\\
}

\begin{document}
\newcommand{\TODO}{\textcolor{red}{TODO}}
\addeditor{nguyen}{NG}{0.7, 0.0, 0.7}
\addeditor{thibault}{TG}{0.0, 0.0, 0.8}
\addeditor{vincent}{VL}{0.0, 0.5, 0.0}
\addeditor{mathieu}{MS}{0.1, 0.5, 0.9}
\addeditor{renaud}{MS}{0.8, 0.4, 0.1}
\showeditstrue
\showeditsfalse


\twocolumn[
\maketitle

\newlength{\teaserheight}
\setlength\teaserheight{1.9cm}
\newlength{\probaheight}

\setlength\probaheight{2.0cm}
\centering
\setlength\lineskip{1.5pt}
\setlength\tabcolsep{1.5pt} 
{\footnotesize
\begin{tabular}{cr}
\begin{tabular}{
>{\centering\arraybackslash}m{\teaserheight}
>{\centering\arraybackslash}m{\teaserheight}
>{\centering\arraybackslash}m{\teaserheight}
>{\centering\arraybackslash}m{\teaserheight}
>{\centering\arraybackslash}m{1.cm}
>{\centering\arraybackslash}m{\teaserheight}
>{\centering\arraybackslash}m{\teaserheight}
>{\centering\arraybackslash}m{\teaserheight}
>{\centering\arraybackslash}m{\teaserheight}
}
 
 Reference & Query & Predicted pose & Pose distribution &
& Reference & Query & Predicted pose & Pose distribution \\
\frame{\includegraphics[height=\teaserheight, ]{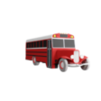}}&
\frame{\includegraphics[height=\teaserheight, ]{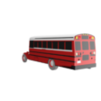}}&
\frame{\includegraphics[height=\teaserheight, ]{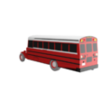}}&
\frame{\includegraphics[height=\teaserheight, ]{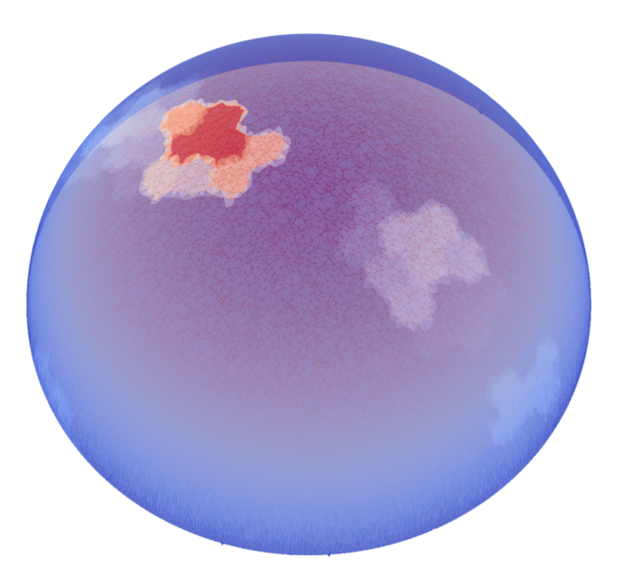}} &
& \frame{\includegraphics[height=\teaserheight, ]{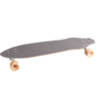}} &
\frame{\includegraphics[height=\teaserheight, ]{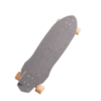}}&
\frame{\includegraphics[height=\teaserheight, ]{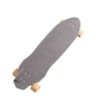}}&
\frame{\includegraphics[height=\teaserheight, ]{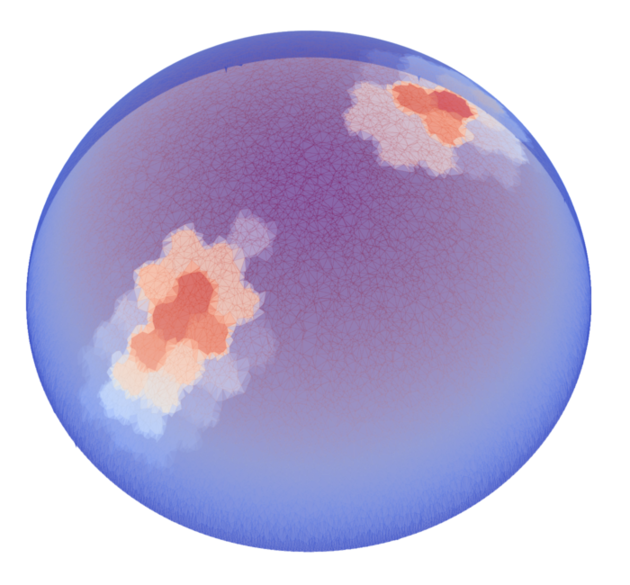}}\\

\frame{\includegraphics[height=\teaserheight, ]{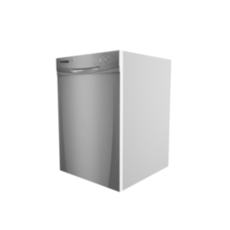}} &
\frame{\includegraphics[height=\teaserheight, ]{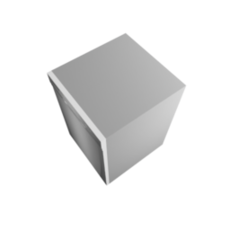}}&
\frame{\includegraphics[height=\teaserheight, ]{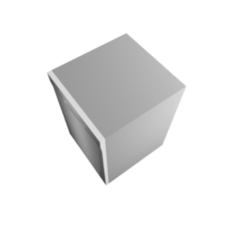}}&
\frame{\includegraphics[height=\teaserheight, ]{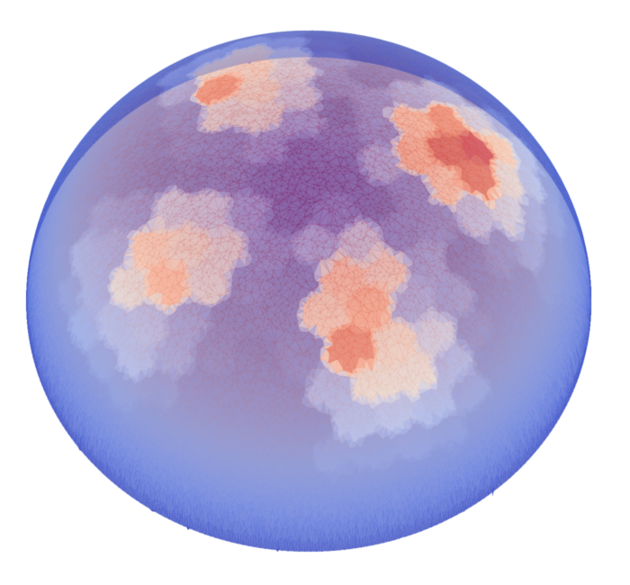}} &

& 
\frame{\includegraphics[height=\teaserheight, ]{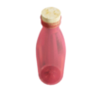}} &
\frame{\includegraphics[height=\teaserheight, ]{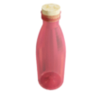}} &
\frame{\includegraphics[height=\teaserheight, ]{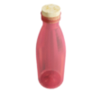}} &
\frame{\includegraphics[height=\teaserheight, ]{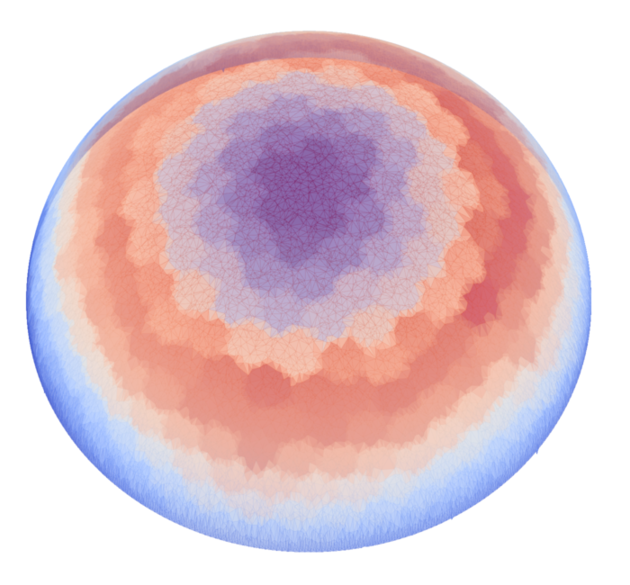}}\\

\frame{\includegraphics[height=\teaserheight, ]{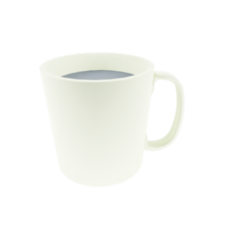}} &
\frame{\includegraphics[height=\teaserheight, ]{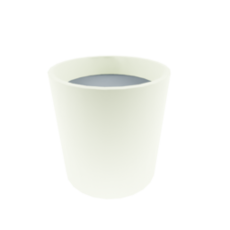}}&
\frame{\includegraphics[height=\teaserheight, ]{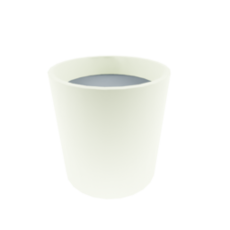}}&
\frame{\includegraphics[height=\teaserheight, ]{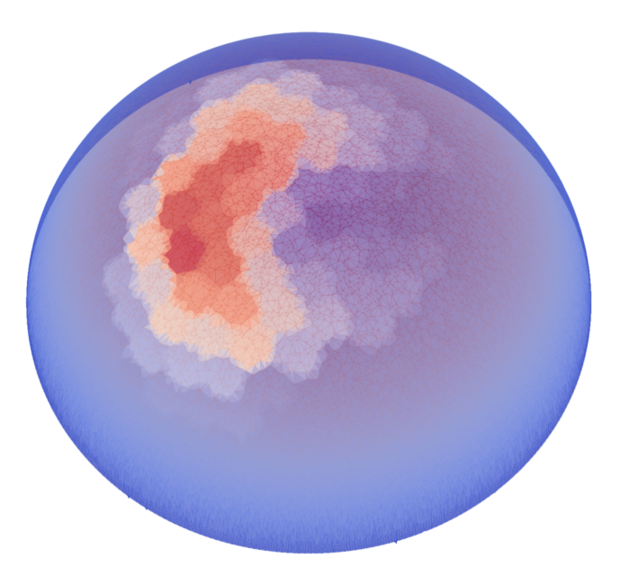}} &
& \frame{\includegraphics[height=\teaserheight, ]{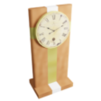}} &
\frame{\includegraphics[height=\teaserheight, ]{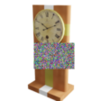}} &
\frame{\includegraphics[height=\teaserheight, ]{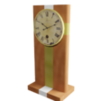}}&
\frame{\includegraphics[height=\teaserheight, ]{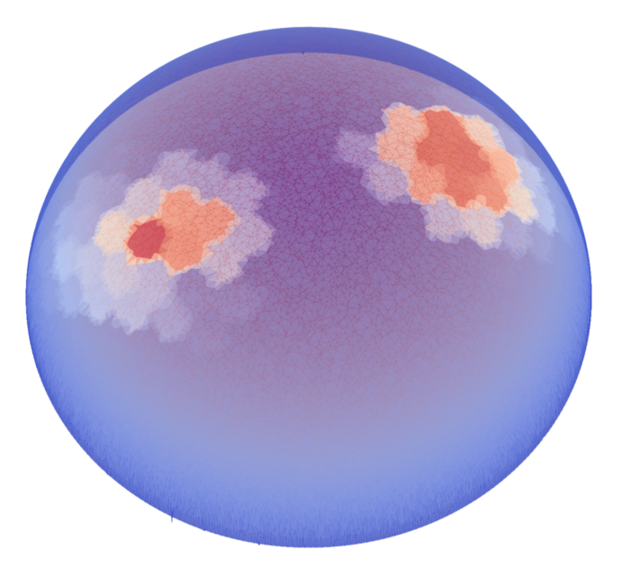}}\\
\end{tabular}

\end{tabular}
}
\vspace{-10pt}
\captionof{figure}{
Given as input a single reference view of a novel object, our method predicts the relative 3D pose (rotation) of a query view and its ambiguities. \textbf{We visualize the predicted pose by rendering the object from this pose, but the 3D model is only used for visualization purposes, not as input to our method.} Our method works by estimating a probability distribution over the space of 3D poses, visualized here on a sphere centered on the object. \textbf{We use the canonical pose of the 3D model to visualize this distribution, but not as input to our method.} From this distribution, we can also identify the pose ambiguities: For example, in the case of the bottle, any pose with the same pitch and roll is possible; in the case of the mug, a range of poses are possible as the handle is not visible in the query image. Our method is also robust to partial occlusions, as shown on the clock hidden in part by a rectangle in the query image.}
\label{fig:teaser}
\vspace{2pt}
\bigbreak]

\begin{abstract}
The practicality of 3D object pose estimation remains limited for many applications due to the need for prior knowledge of a 3D model and a training period for new objects. To address this limitation, we propose an approach that takes a single image of a new object as input and predicts the relative pose of this object in new images without prior knowledge of the object's 3D model and without requiring training time for new objects and categories. We achieve this by training a model to directly predict discriminative embeddings for viewpoints surrounding the object. This prediction is done using a simple U-Net architecture with attention and conditioned on the desired pose, which yields extremely fast inference. We compare our approach to state-of-the-art methods and show it outperforms them both in terms of accuracy and robustness. 
\vspace{0.2cm}
\end{abstract}
\thispagestyle{plain}
\pagestyle{plain}

{
\setstretch{0.95}

\section{Introduction}
\label{sec:introduction}
Estimating the 3D pose of objects has seen significant progress in the past decade with regard to both robustness and accuracy~\cite{kehl-iccv17-ssd6d,rad-iccv17-bb8,tekin-cvpr18-realtimeseamlesssingleshot,li-eccv18-deepim,zakharov-iccv19-dpod}. Specifically, there has been a considerable increase in robustness to partial occlusions~\cite{peng-cvpr19-pvnet,hu-cvpr19-segmentationdriven6dobjectposeestimation,oberweger-eccv18-makingdeepheatmapsrobust}, and the need for large amounts of real annotated training images has been relaxed through the use of domain transfer~\cite{baek-cvpr20-weaklysuperviseddomainadaptation}, domain randomization~\cite{tremblay-18-deepobjectposeestimation,loing-ijcv18-virtual,labbe-eccv20-cosypose,sundermeyer-cvpr20-multipathlearning}, and self-supervised learning techniques~\cite{sundermeyer-ijcv20-augmentedautoencoders} that leverage synthetic images for training.

Unfortunately, the practicality of 3D object pose estimation remains limited for many applications, including robotics and augmented reality. Typically, existing approaches require a 3D model~\cite{xiao-bmvc19-posefromshape,megapose,nguyen2022templates,nguyen2024gigaPose}, a video sequence~\cite{onepose,oneposepp}, or sparse multiple images of the target object~\cite{relpose}, and a training stage. Several techniques aim to prevent the need for retraining by assuming that new objects fall into a recognized category~\cite{Grabner_CVPR18,Wang_2019_NOCS}, share similarities with the previously trained examples as in the T-LESS dataset~\cite{sundermeyer-cvpr20-multipathlearning}, or exhibit noticeable corners~\cite{Pitteri20203DOD}.

In this paper, we introduce an approach, which we call {\bf NOPE} for {\bf N}ovel {\bf O}bject {\bf P}ose {\bf E}stimation, that only requires a single image of the new object to predict the relative pose of this object in any new images, without the need for the object's 3D model and without training on the new object. This is a very challenging task, as, by contrast with the multiple views used in \cite{onepose,relpose} for example, a single view only provides limited information about the object's geometry.

\vincent{
To achieve this, we train NOPE to predict the appearance of the object under novel views. We use these predictions as `templates' annotated with the corresponding poses.  Matching these templates with new input views lets us estimate the object relative pose with respect to the initial view. This approach is motivated by the good performance of recent related work~\cite{nguyen2022templates, shugurov_osop_2022}. In particular, \cite{nguyen2022templates} showed that template matching can be extremely fast and robust to partial occlusions. This contrasts with methods that rely on a deep network to predict the probability of a pose~\cite{relpose}.

Since our method relies on predicting the appearance of the target object, it relates to recent developments in novel view synthesis. However, it has two critical differences: The first difference is that instead of predicting color images, we directly predict discriminative embeddings of the views. These embeddings are extracted by passing the input image through a U-Net architecture with attention and conditioned on the desired pose for the new view. 

The second main difference of our approach with novel view synthesis is more fundamental. We first note that generating novel views given a single view of an object is ambiguous. Novel view synthesis usually focuses on generating a single possible image for a given point of view. This is however not suitable for our purpose: The view synthesis method will ``invent'' the parts that were not visible in the input view. As illustrated in Figure~\ref{fig:motivation}, these invented parts create a plausible novel view but there is no guarantee this view actually corresponds to the actual view. For our goal of pose estimation, the invented parts will not match in general the query view and this will result in incorrect pose estimation. The limitations of using novel view synthesis for pose estimation will further be quantitatively demonstrated in our experiments~(see Table~\ref{tab:shapeNet}).

Our approach to handling the ambiguities in novel view synthesis for template matching is to consider the \emph{distribution} of all the possible appearances of the object for the target viewpoint. More exactly, we train NOPE to predict the average of all the possible appearances of the object. We then treat the predicted average as a template: Under some simple assumptions, the distance between this template and the query view is directly related to the probability of the query view to be a sample from the distribution of the possible appearances of the object. This approach allows us to deal with the ambiguities of novel view prediction in a robust and efficient way:  Predicting the average views is just a direct inference of NOPE and is thus very fast,  and robust to partial occlusions thank to template-matching. 

Furthermore, our approach can identify the pose ambiguities due, for example, to symmetries~\cite{tombari}, even if we do not have access to the object 3D model but only to a single view. To this end, we estimate the distribution over all poses for the query, which becomes increasingly less peaked as the pose suffers from increasingly many ambiguities. Figure~\ref{fig:teaser} depicts a variety of ambiguous and unambiguous cases with their pose distributions.
}



\begin{figure}[t]
\newlength{\plotheight}
\setlength\plotheight{1.5cm}
\centering
\setlength\lineskip{2.0pt}
\setlength\tabcolsep{2.0pt} 
{\small
\begin{tabular}{
>{\centering\arraybackslash}m{\plotheight}
>{\centering\arraybackslash}m{\plotheight}
>{\centering\arraybackslash}m{\plotheight}
>{\centering\arraybackslash}m{\plotheight}
>{\centering\arraybackslash}m{\plotheight}
}
 & & Generated & Recovered &  \\
 & & view from & pose by & Estimated \\
 & & the query & template &  pose \\
Reference & Query & GT pose & matching & distribution \\
\frame{\includegraphics[height=\plotheight, ]{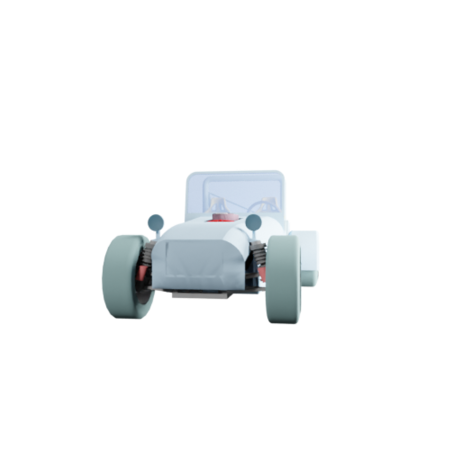}}&
\frame{\includegraphics[height=\plotheight, ]{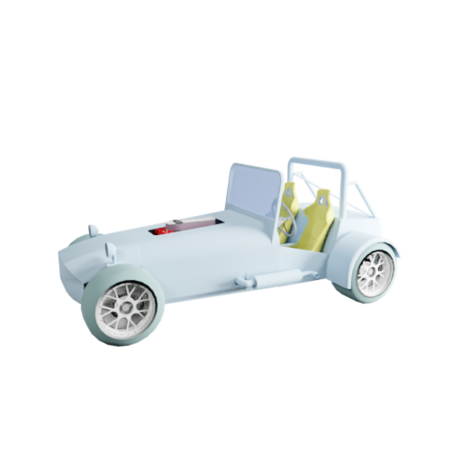}}&
\frame{\includegraphics[height=\plotheight, ]{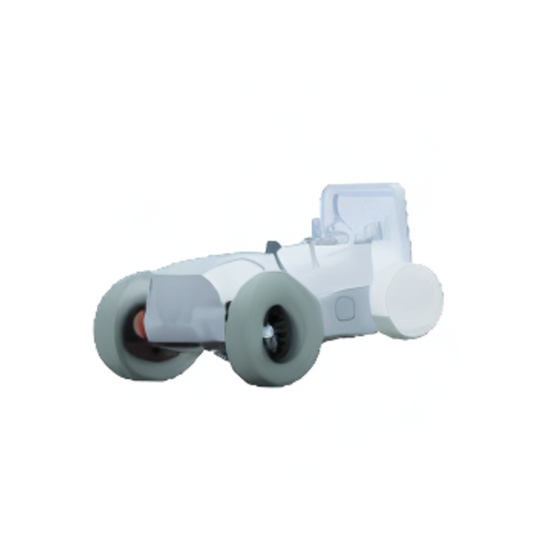}}&
\frame{\includegraphics[height=\plotheight, ]{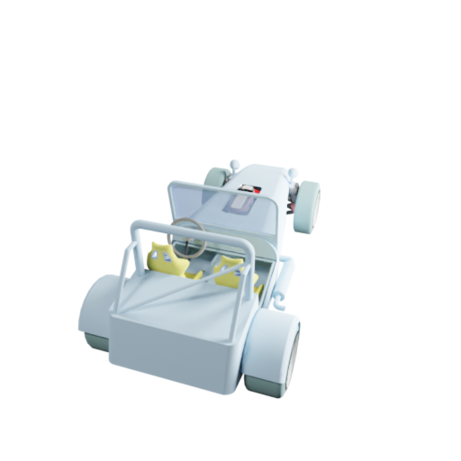}}&
\frame{\includegraphics[height=\plotheight, ]{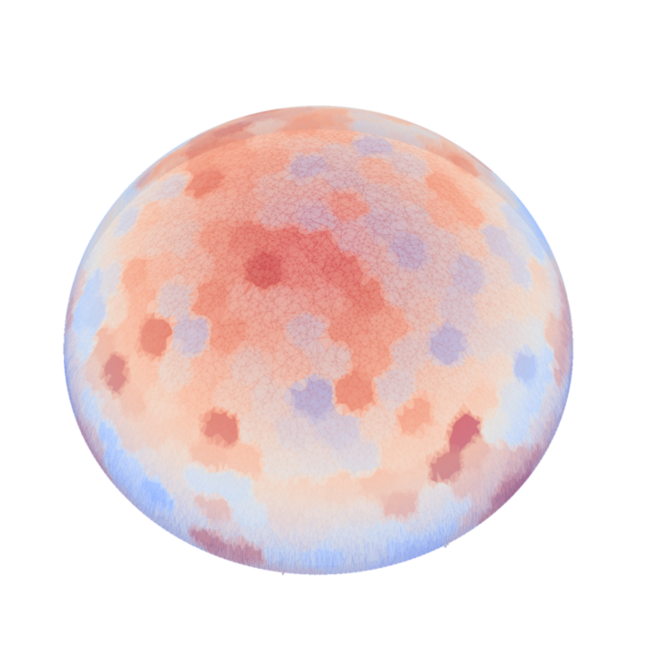}}\\

\frame{\includegraphics[height=\plotheight, ]{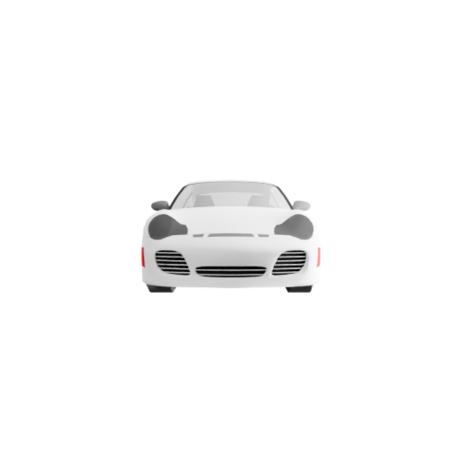}}&
\frame{\includegraphics[height=\plotheight, ]{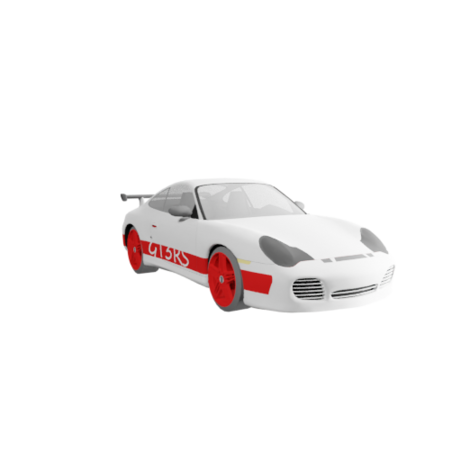}}&
\frame{\includegraphics[height=\plotheight, ]{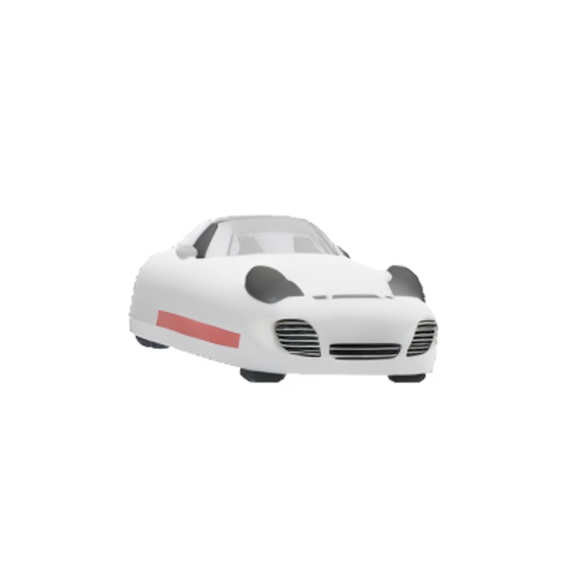}}&
\frame{\includegraphics[height=\plotheight, ]{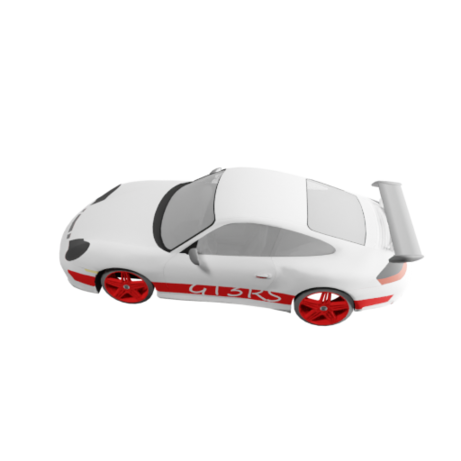}}&
\frame{\includegraphics[height=\plotheight, ]{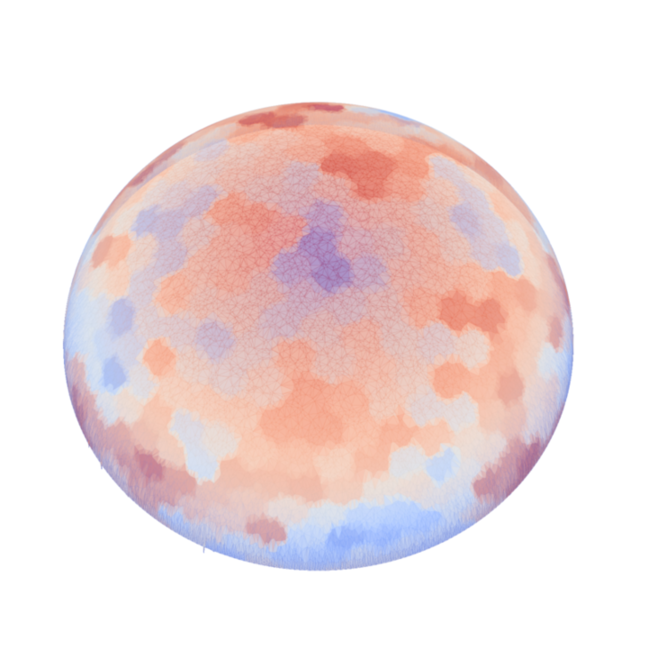}}\\
\end{tabular}
}
\vspace*{-10pt}
\caption{
\vincent{
\textbf{The limit of novel view synthesis for pose prediction.} While the images generated by Wonder3D~\cite{long2023wonder3d} \vincentrmk{Can we have Wonder3D here?} \nguyenrmk{I put two samples of Wonder3d} look very realistic, they have to invent unseen parts, impairing the similarity computation between the query image and the generated view, and hence the pose estimation: The probability distributions computed by template matching do not peak on the right pose but show many wrong local maxima. This is not a limitation of Wonder3D but of view synthesis from a single view in general.
}
}
\label{fig:motivation}
\end{figure}


In summary, our main contribution is to show we can  efficiently and reliably recover the relative pose of an unseen object in novel views given only a single view of that object \renaud{as reference}. To the best of our knowledge, our approach is the first to predict ambiguities due to symmetries and partial occlusions of unseen objects from only a single view.


\section{Related Work}
\label{sec:relatedwork}
In this section, we first review various approaches to novel view synthesis. We then shift our focus to pose estimation techniques that aim to achieve generalization.

\subsection{Novel view synthesis from a single image}

Our method generates discriminative feature views, which are conditioned on a reference view and the relative pose between the views. This relates to the pioneering work of NeRFs~\cite{mildenhall2020nerf} since it performs novel-view synthesis. While recent advancements have improved the speed of NeRFs~\cite{muller_instant_2022, sparsevoxelgrid, plenoxel}, our approach is still orders of magnitude faster as it does not require the creation of a full 3D volumetric model. Furthermore, our approach only requires a single input view, whereas a typical NeRF setup necessitates around 50 views. Reducing the number of views required for NeRF reconstruction remains an active research area, especially in the single-view scenario~\cite{pixelNerf, realfusion}.

Recent works~\cite{sparsefusion, realfusion}  have had successes generating novel views via NERFs  using a sparse set of views as input by leveraging 2D diffusion models. For images, the breakthrough in diffusion models~\cite{ddpm,song2020denoising} have unlocked several workflows~\cite{ramesh10hierarchical, saharia2022photorealistic,ruiz2022dreambooth}. For 3D applications, DreamFusion~\cite{poole2022dreamfusion} pioneered a score-distillation sampling that allows for the use of a 2D diffusion model as an image-based loss, leveraged by 3D applications via differentiable rendering. This has resulted in significant improvements for tasks previously trained with a CLIP-based image loss~\cite{clip, dreamfields, CLIPmesh, magic3d, Clipasso, alainjay}.  By building on top of score-distillation sampling, SparseFusion~\cite{sparsefusion} reconstructs a NeRF scene with as few as two views with relative pose, while the concurrent work RealFusion~\cite{realfusion} does it from a single input view, although the reconstruction time is impractical for real-time applications. Our approach is much faster as we do not create a 3D representation of the object.

Closest to us, 3DiM~\cite{3dim} and Zero-1-to-3~\cite{liu2023zero} generate novel views of an object by conditioning a diffusion model on the pose. Instead of leveraging foundation diffusion models in 2D like DreamFusion~\cite{poole2022dreamfusion} does, they retrain a diffusion model specifically for this task. While they have not applied their approach to template-based pose estimation, we design such a baseline and compare against it. We find that the diffusion model tends to change the texture or invent wrong details which hinders the performance of the template-based approach. In contrast, our approach generates average novel views directly in an embedding space instead of a pixel space, which is much more efficient~\cite{nguyen2022templates}.  

Finally, several methods~\cite{mariotti_semi-supervised_2020,mariotti_viewnerf_2022} generate novel views by conditioning a feed-forward neural network on the 3D pose, which we also do with a U-Net. We share with these methods an advantage in speed: such feed-forward neural network are one or two orders of magnitude faster than current diffusion models. However, the way we perform pose estimation is fundamentally different. We use novel-view synthesis in a template-based matching approach~\cite{nguyen2022templates}, while they use it in a regression-based optimization. In practice, we found these methods to work well on a limited number of object categories, and we observed their performance to deteriorate significantly when \nguyen{testing} on novel categories.

\vspace{-4pt}
\subsection{Generalizable object pose estimation}
\vspace{-5pt}

Several techniques have been explored to generalize better to unseen object pose estimation, such as generic 2D-3D correspondences~\cite{Pitteri20203DOD}, an energy-based strategy~\cite{relpose}, keypoint matching~\cite{onepose}, or template matching~\cite{nguyen2022templates, shugurov_osop_2022, chen_fusion, liu2022gen6d, megapose, nguyen2024gigaPose}. Despite significant progress, these methods either need an accurate 3D model of the target or they  rely on multiple annotated reference images from different viewpoints. These 3D annotations are challenging to obtain in practice. By contrast, we propose a strategy that works with neither the 3D model of the target nor the annotation of multiple views. More importantly, our method predicts accurate poses with only a single reference image, and generalizes to novel objects without retraining.

\section{Method}
\label{sec:method}
\vspace{-6pt}
\vincent{
In this section, we first introduce our formalism, then describe our architecture and how we train it, and finally how we use it for pose prediction and for identifying pose ambiguities.
}

\begin{figure*}[!t]
    \begin{center}
    \includegraphics[width=1\linewidth]{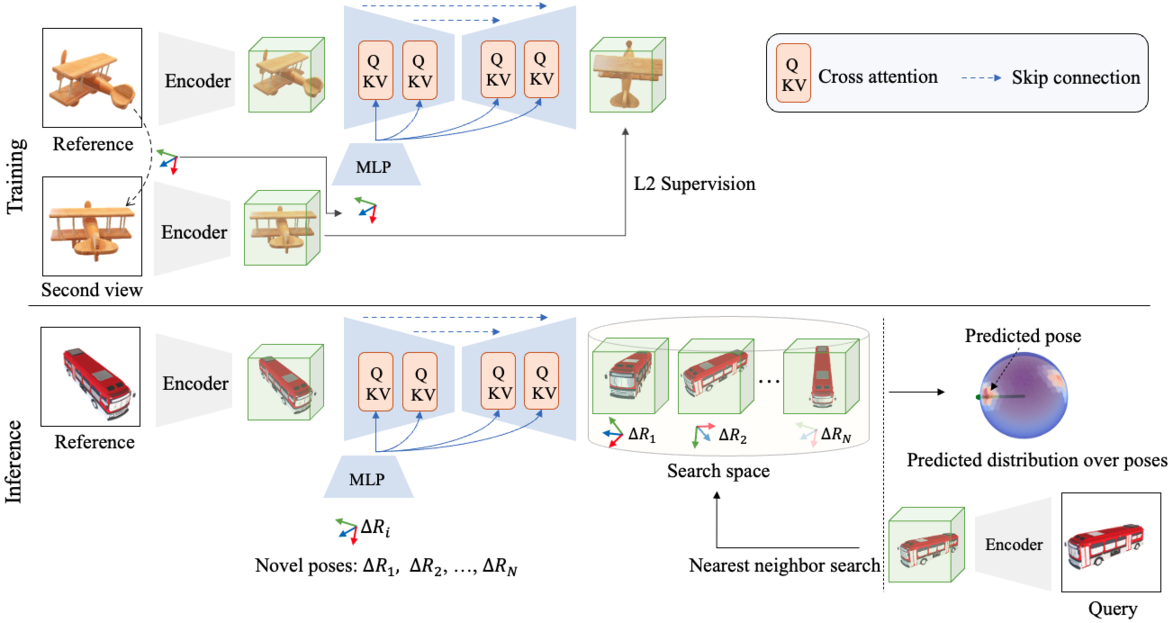}
    \end{center}
    \vspace{-20pt}
    \caption{
        \label{fig:framework}
        {\bf Overview.} During training, we train a U-Net to predict the embedding of a novel view of an object, given a reference image of the object and a relative pose. The U-Net is conditioned on an embedding of the relative pose computed using an MLP, which we train jointly with the U-Net. 
        At inference, our method first takes as input a reference image of a new object and predicts the embeddings of views of the object under many relative poses. This inference takes around 1 second on a single GPU V100.
        Then, given a query image of the object, we first compute its embedding and match it against the set of predicted embeddings.  This gives us a distribution over the possible relative poses between the reference and query images, where the maximum corresponds to the predicted pose. 
        \vspace{-5pt} 
        }
\end{figure*}


\subsection{Formalization}

Given a reference image $I_r$ of a target object and a query image $I_q$ of the same object, we would like to estimate the probability $p(\Delta R \;|\; I_r, I_q)$ that the relative motion between $I_r$ and $I_q$ is a certain discretized relative pose $\Delta R$. We assume that this probability follows a normal distribution in the embedding space of the images:
\begin{equation}
p(\Delta R \;|\; I_r, I_q) = \mathcal{N}(\rve_q \;|\; \rve(\rve_r, \Delta R), \Sigma(\rve_r, \Delta R)) \> ,
\label{eq:N}
\end{equation}
where $\rve_q$ and $\rve_r$ are the embeddings for query image $I_q$ and reference image $I_r$ respectively, $\rve(\rve_r, \Delta R)$ is the mean of the normal distribution, and $\Sigma(\rve_r, \Delta R)$ its covariance. This approach allows us to handle the fact that the object can have various appearances from viewpoint $\Delta R$ given the reference image, as discussed in the introduction.

We take the mean $\rve(\rve_r, \Delta R)$ as the average embedding for the appearance of the object from pose $\Delta R$ over the possible 3D shapes for the object:
\begin{equation}
\rve(\rve_r, \Delta R) = \int_{\mathcal{M}} \rve(\Delta R, \mathcal{M}) p(\mathcal{M} | \rve_r)d\mathcal{M} \> ,
\end{equation}
with $\mathcal{M}$ a 3D model of testing object and $\rve(\Delta R, \mathcal{M})$ the image embedding of same object under pose $\Delta R$. $\rve(\rve_r, \Delta R)$ may look complicated to compute, but it is in fact easy to train a deep network to predict it using the L2 loss:
\begin{equation}
\sum_{(\rve_1, \rve_2, \Delta R)} \|F(\rve_r, \Delta R) - \rve_2 \|^2 \> .
\end{equation}
$F$ denotes the network, $(\rve_1, \rve_2, \Delta R)$ is a training sample where $\rve_1$ is the embedding for a view of a training object and $\rve_2$ the embedding for the view of the same object after pose change $\Delta R$. During training, given enough  samples, $F(\rve_r, \Delta R)$ will converge naturally towards $\rve(\rve_r, \Delta R)$.

\subsection{Framework}
\label{sec:framework}

Figure~\ref{fig:framework} gives an overview of our approach. 
We train a deep architecture to predict the average embeddings of novel views of an object using pairs of images of objects and the corresponding pose changes from a first set of object categories. In practice, we consider embeddings computed from the pretrained VAE of~\cite{stable-diffusion}, as it was shown to be robust for template matching. To generate these embeddings, we use a U-Net-like network with a pose conditioning mechanism that is very close to the one of 3DiM~\cite{3dim}. 

More precisely, we first use an MLP to convert the desired relative viewpoint $\Delta R$ with respect to the object pose in the reference view to a pose embedding. We then integrate this pose embedding into the feature map at every stage of our U-Net using cross-attention, as  in \cite{stable-diffusion}. 

\vspace*{-15pt}
\paragraph{Training.} At each iteration, we build a batch composed of $N$ pairs of images, a reference image and another image of the same object with a known relative pose. The U-Net model takes as input the embedding of the reference image and as conditioning the embedding of the relative pose to predict an embedding for the second image. We jointly optimize the U-Net and the MLP by minimizing the Euclidean distance between this predicted embedding and the embedding of the query image. Note that we freeze the pretrained VAE network of \cite{stable-diffusion} during the training.

By training it on a dataset of diverse objects, this architecture generalizes well to  novel unseen object categories. Interestingly, our method does not explicitly learn any symmetries during training, but it is able to detect pose ambiguities during testing as discussed below.

\subsection{Pose prediction}
\label{sec:templateMatching}

\vspace{-4pt}
\paragraph{Template matching.} 
Once our architecture is trained, we can use it to generate the embeddings for novel views: Given a reference image $I_r$ and a set of $N$ relative viewpoints $\mathcal{P}=\left(\relativeR_1, \relativeR_2, \dots, \relativeR_N \right)$, we can obtain a corresponding set of predicted embeddings $(\mathbf{e}_1, \mathbf{e}_2, \dots, \mathbf{e}_N)$. To define these viewpoints, we follow the approach used in \cite{nguyen2022templates}: We start with a regular icosahedron and subdivide each triangle recursively into four smaller triangles twice to get 342 final viewpoints. Finally, we simply perform a nearest neighbor search to determine the reference point that has the embedding closest to the embedding of the query image. 

\vspace{-8pt}
\paragraph{Detecting pose ambiguities.} Pose ambiguities arise when the object has symmetries or when an object part that could remove the ambiguity is not visible, as for the mug in Figure~\ref{fig:teaser}. By considering the distance between the embedding of the query image and the generated embeddings, we not only can predict a single pose but also identify all the other poses that are possible given the reference and query views.

\newcommand{\be}{{\bf e}}

This can be done simply by relying on the normal distribution introduced in Eq.~(\ref{eq:N}):
\begin{equation}
\log p(\Delta R \; | \; I_r, I_q) \propto \|F(\rve_r, \Delta R) - \rve_q\|^2 \> .
\end{equation}


\begin{figure}[t]
\newlength{\plotHeight}
\setlength\plotHeight{1.8cm}
\centering
\setlength\lineskip{1.5pt}
\setlength\tabcolsep{1.5pt} 
{\footnotesize
\begin{tabular}{
>{\centering\arraybackslash}m{\plotHeight}
>{\centering\arraybackslash}m{\plotHeight}
>{\centering\arraybackslash}m{\plotHeight}
>{\centering\arraybackslash}m{\plotHeight}
}
No symmetry & 90-symmetry &  180-symmetry & Circular symmetry \\ 
\frame{\includegraphics[width=\plotHeight, ]{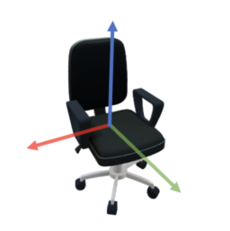}}&
\frame{\includegraphics[width=\plotHeight, ]{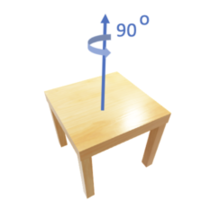}}&
\frame{\includegraphics[width=\plotHeight, ]{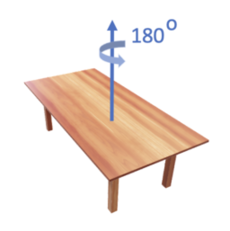}}&
\frame{\includegraphics[width=\plotHeight, ]{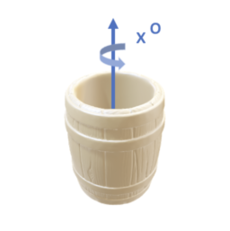}}\\
\frame{\includegraphics[width=\plotHeight, ]{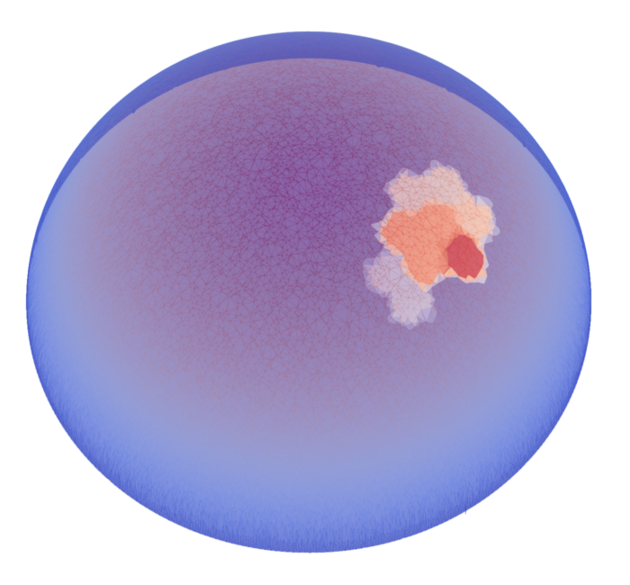}}&
\frame{\includegraphics[width=\plotHeight, ]{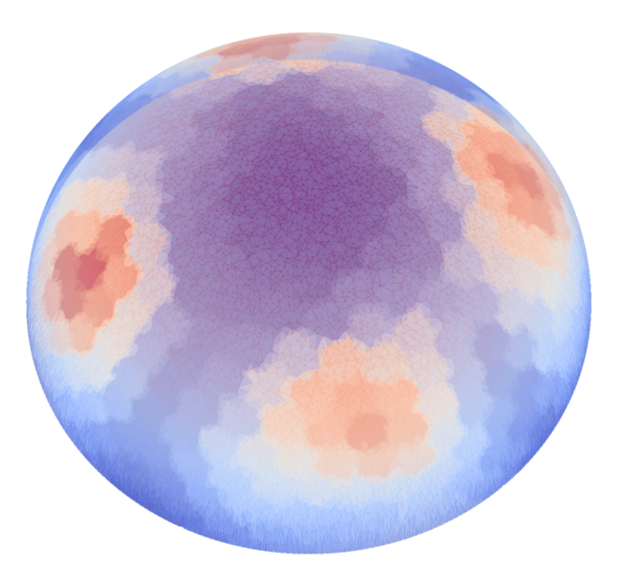}}&
\frame{\includegraphics[width=\plotHeight, ]{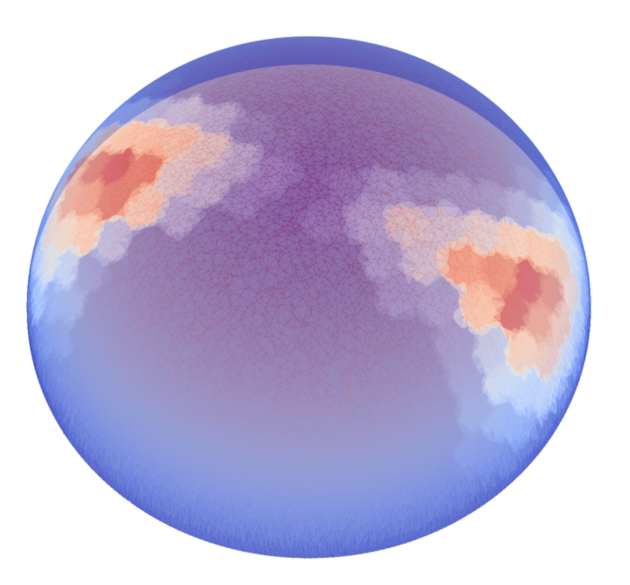}}&
\frame{\includegraphics[width=\plotHeight, ]{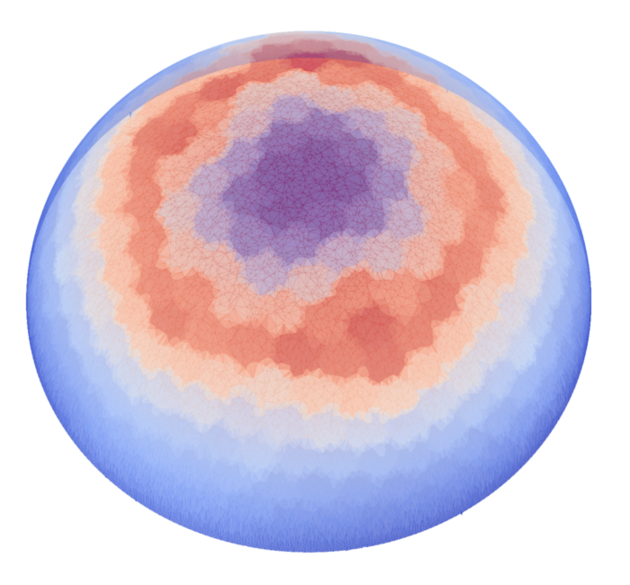}}\\
\end{tabular}
}
\caption{{\bf Object symmetries and the pose ambiguities they may generate}, as estimated by our method given a pair of reference and query images.}
\label{fig:symmetry}
\end{figure} 

To illustrate this, we show in Figure~\ref{fig:symmetry} three distinct types of symmetry and visualize the pose distribution for corresponding pairs of  reference and query images~(not shown). The number of regions with high similarity scores is consistent with the number of symmetries and pose ambiguities:  If an object has no symmetry, the probability distribution has a clear mode. The probability distribution for objects with symmetries have typically several modes or even a continuous high-probability region in case of rotational symmetry. We provide additional qualitative results in Section~\ref{sec:experiments}.

\section{Experiments}
\label{sec:experiments}
\vspace{-6pt}

In this section, we first describe our experimental setup in Section~\ref{sec:experimental setup}. We then compare our method to others \cite{nguyen_pizza_2022, mariotti_semi-supervised_2020, mariotti_viewnet_2021, 3dim, sundermeyer-cvpr20-multipathlearning, nguyen2022templates} on both synthetic and real-world datasets in Section~\ref{sec:main_results}. Section~\ref{sec:occlusions} reports an evaluation of the robustness to partial occlusions. We  provide the run-time in Section~\ref{sec:run_time}. Finally, we discuss  failure cases in Section~\ref{sec:failureCases}. An ablation study is provided in the supp.~mat.


\subsection{Experimental setup}
\label{sec:experimental setup}

\begin{figure*}[t]
\setlength\plotHeight{1.35cm}
\centering
\setlength\lineskip{0.5pt}
\setlength\tabcolsep{0.0pt} 
{\footnotesize
\begin{tabular}{c}
\begin{tabular}{
>{\centering\arraybackslash}m{\plotHeight}
>{\centering\arraybackslash}m{\plotHeight}
>{\centering\arraybackslash}m{\plotHeight}
>{\centering\arraybackslash}m{\plotHeight}
>{\centering\arraybackslash}m{\plotHeight}
>{\centering\arraybackslash}m{\plotHeight}
>{\centering\arraybackslash}m{\plotHeight}
>{\centering\arraybackslash}m{\plotHeight}
>{\centering\arraybackslash}m{\plotHeight}
>{\centering\arraybackslash}m{\plotHeight}
>{\centering\arraybackslash}m{\plotHeight}
>{\centering\arraybackslash}m{\plotHeight}
>{\centering\arraybackslash}m{\plotHeight}
}
\toprule
\multicolumn{13}{c}{\textbf{Object categories in the training set}}\\
airplane & bench & cabinet & car & chair & display & lamp &  loudspeaker & rifle & sofa & table & telephone & vessel\\
\midrule
\includegraphics[height=\plotHeight, ]{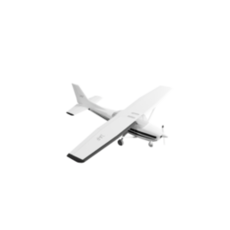}&
\includegraphics[height=\plotHeight, ]{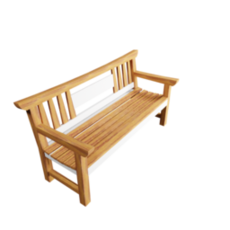}&
\includegraphics[height=\plotHeight, ]{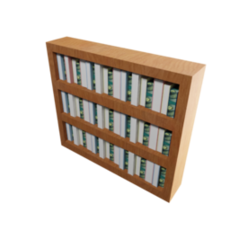}&
\includegraphics[height=\plotHeight, ]{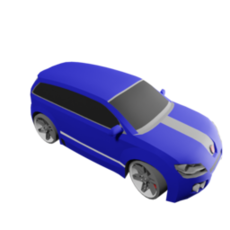}&
\includegraphics[height=\plotHeight, ]{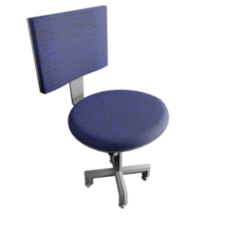}&
\includegraphics[height=\plotHeight, ]{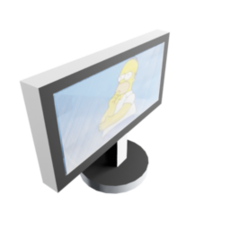}&
\includegraphics[height=\plotHeight, ]{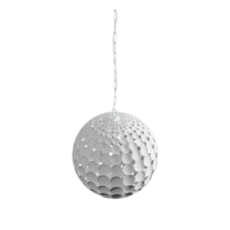}&
\includegraphics[height=\plotHeight, ]{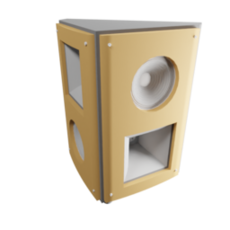}&
\includegraphics[height=\plotHeight, ]{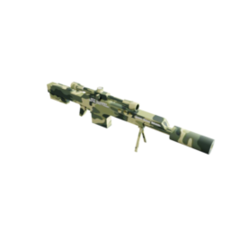}&
\includegraphics[height=\plotHeight, ]{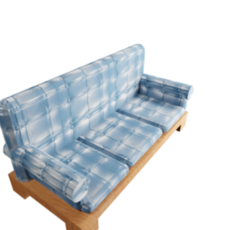}&
\includegraphics[height=\plotHeight, ]{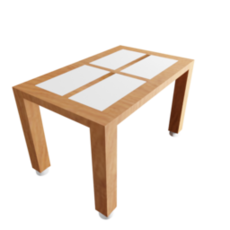}&
\includegraphics[height=\plotHeight, ]{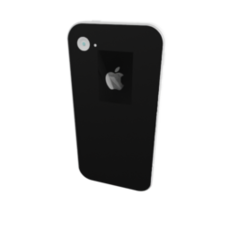}&
\includegraphics[height=\plotHeight, ]{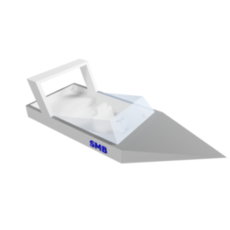}\\
\bottomrule
\end{tabular}\\[1.2cm]

\begin{tabular}{
>{\centering\arraybackslash}m{\plotHeight}
>{\centering\arraybackslash}m{\plotHeight}
>{\centering\arraybackslash}m{\plotHeight}
>{\centering\arraybackslash}m{\plotHeight}
>{\centering\arraybackslash}m{\plotHeight}
>{\centering\arraybackslash}m{\plotHeight}
>{\centering\arraybackslash}m{\plotHeight}
>{\centering\arraybackslash}m{\plotHeight}
>{\centering\arraybackslash}m{\plotHeight}
>{\centering\arraybackslash}m{\plotHeight}
}
\toprule
\multicolumn{10}{c}{\textbf{Object categories in the test set}}\\
bottle  & bus & clock & dishwasher & guitar & mug & pistol & skateboard & train & washer\\
\midrule
\includegraphics[height=\plotHeight, ]{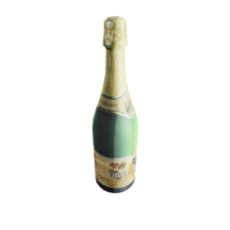}&
\includegraphics[height=\plotHeight, ]{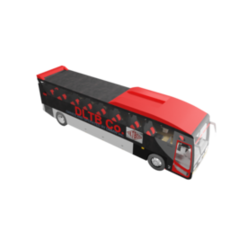}&
\includegraphics[height=\plotHeight, ]{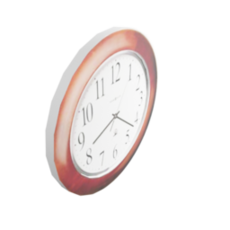}&
\includegraphics[height=\plotHeight, ]{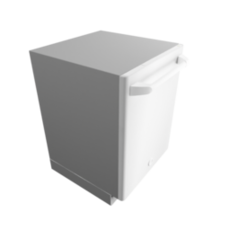}&
\includegraphics[height=\plotHeight, ]{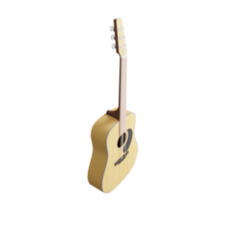}&
\includegraphics[height=\plotHeight, ]{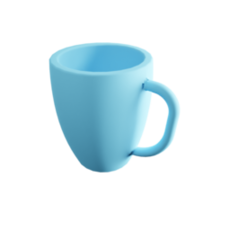}&
\includegraphics[height=\plotHeight, ]{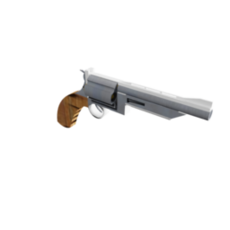}&
\includegraphics[height=\plotHeight, ]{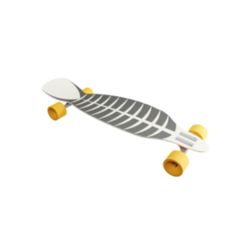}&
\includegraphics[height=\plotHeight, ]{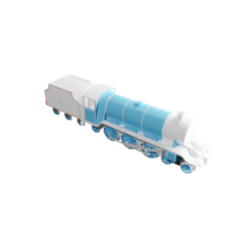}&
\includegraphics[height=\plotHeight, ]{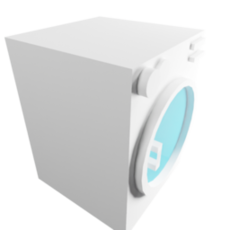}\\
\bottomrule
\end{tabular}\\
\end{tabular}}
\caption{{\bf \nguyen{Visualization of training and test sets from the ShapeNet dataset~\cite{chang2015shapenet}}.} 
The shapes and appearances in the training and test sets are very different and thus constitute a good test bed for generalization to unseen categories.
}
\label{fig:shapeNet}
\end{figure*}

\begin{table*}[!t]
\addtolength{\tabcolsep}{1pt}
\definecolor{light_yellow}{RGB}{255, 244, 200}
\definecolor{orange}{RGB}{255, 205, 115}
\centering
    \scalebox{.87}{
    \begin{tabular}{@{}l l | ccccccccccc | c@{}}
	\toprule
	& Method &  novel inst. & bottle$^*$ & bus & clock & dishwasher & guitar & mug & pistol & skateboard & train & washer & mean \\
	\midrule
	\parbox[t]{2mm}{\multirow{7}{*}{\rotatebox[origin=c]{90}{$\bm{\AccThirty}\uparrow$}}} 
	& ViewNet \cite{mariotti_viewnet_2021}  & \bf 77.5 & 48.4 & 36.2 & 23.5 & 16.4 & 37.8 & 31.3 & 17.9 & 33.9 & 44.8 & 25.1 & 35.7  \\
	& SSVE \cite{mariotti_semi-supervised_2020} & 75.3 & 61.5 & 38.2 & 41.8 & 21.3 & 46.8 & 38.4 & 36.8 & \bf 62.3 & 41.5 & 50.8 & 46.8  \\
	& PIZZA \cite{nguyen_pizza_2022} & 72.3 & 76.0 & 38.6 & 38.5 & 32.6 & 30.8 & 35.6 & 40.4 & 58.3 & 52.9 & \bf 61.0 & 48.8  \\
	& 3DiM \cite{3dim} & 77.3 & 95.1 & 43.5 & 23.6 & 24.5 & 36.0 & 32.0 & 31.9 & 50.3 & 37.0 & 56.1 & 46.1	\\
	& Ours (top 1)  & 75.5 & \bf 96.0 & \bf 53.6 & \bf 48.0 & \bf 48.0 & \bf 49.0 & \bf 44.6 & \bf 69.0 & 57.8 & \bf 55.2 & 60.6 & \bf 59.8 \\
	\cmidrule{2-14}
	& Ours (top 3) & 92.0 & 97.4 & 83.8 & 73.4 & 78.5 & 66.8 & 56.0 & 83.8 & 86.2 & 86.0 & 84.4 & 80.8 \\
	& Ours (top 5) & 95.5 & 97.8 & 89.8 & 80.4 & 88.2 & 74.6 & 62.8 & 88.4 & 92.8 & 95.4 & 93.4 & 87.1\\
    \midrule
    
    \parbox[t]{2mm}{\multirow{7}{*}{\rotatebox[origin=c]{90}{$\bm{\MedErr}\downarrow$}}}  & ViewNet \cite{mariotti_viewnet_2021} & $\phantom{0}$6.6 & 26.7 & 35.8 & 40.3 & 96.3 & 50.6 & 51.6 & 42.8 & 37.4 & 26.8 & 44.3 & 41.7 \\
    
    & SSVE \cite{mariotti_semi-supervised_2020} & $\phantom{0}$6.1 & 23.8 & 45.2 & 41.9 & 90.4 & 47.6 & 49.6 & 24.0 & 13.5 & 24.9 & 48.1 & 37.7  \\
    
    & PIZZA     \cite{nguyen_pizza_2022} & $\phantom{0}$5.8 & 25.5 & 26.4 & 43.2 & 80.6 & 40.2 & 45.5 & 23.4 & 17.3 & 20.3 & 38.5 & 33.3 \\
    
    & 3DiM \cite{3dim} & \bf $\phantom{0}$5.7 & \bf $\phantom{0}$1.8 & 19.8 & 47.3 &  98.8 & 35.2 & 35.7 & 21.2 & \bf 12.5 & \bf 17.6 & 19.2 & 28.6 \\
    
	& Ours (top 1) & $\phantom{0}$8.1 & \bf $\phantom{0}$1.8 & \bf 18.4 & \bf 39.9 & \bf 77.6 & \bf 31.6 & \bf 35.5 & \bf 13.4 & 15.5 &  18.3 & \bf $\phantom{0}$8.5 & \bf 24.4 \\
 
	\cmidrule{2-14}
	& Ours (top 3) & $\phantom{0}$5.0 & $\phantom{0}$1.3 & $\phantom{0}$5.8 & $\phantom{0}$9.1 & $\phantom{0}$4.8 & 16.0 & 22.6 & $\phantom{0}$8.1 & $\phantom{0}$6.5 & $\phantom{0}$6.7 & $\phantom{0}$5.7  & $\phantom{0}$8.3 \\
	& Ours (top 5) & $\phantom{0}$4.5 & $\phantom{0}$1.2 & $\phantom{0}$4.5 & $\phantom{0}$7.1 & $\phantom{0}$4.4 & 11.6 & 18.4 & $\phantom{0}$6.1 & $\phantom{0}$5.6 & $\phantom{0}$4.9 & $\phantom{0}$5.0 & $\phantom{0}$6.6 \\
	\bottomrule
	\end{tabular}
    }
    \caption{{\bf \nguyen{Quantitative results on ShapeNet}.} *We treat “bottle" as a symmetric category, i.e., the error is only the difference of elevation angle. Since the quality of prediction may depend on the reference image, we report the score as the average over 5 runs with 5 different reference images.  
    }
  \label{tab:shapeNet}
  \end{table*} 
To the best of our knowledge, we are the first method addressing the problem of object pose estimation from a single image when the object belongs to a category not seen during training: PIZZA~\cite{nguyen_pizza_2022} evaluated on the DeepIM refinement benchmark, which is made of pairs of images with a small relative pose; SSVE~\cite{mariotti_semi-supervised_2020} and ViewNet~\cite{mariotti_viewnet_2021} evaluated only on objects from categories seen during training.  We therefore  had to create a new benchmark to evaluate our method. 

\vspace{-12pt}
\paragraph{Synthetic dataset.} 
We created a dataset as in FORGE~\cite{forge_jiang} using the same ShapeNet~\cite{chang2015shapenet} object categories.
For the training set, we randomly select 1000 object instances from each of the 13 categories as done in FORGE~(\textit{airplane, bench, cabinet, car, chair, display, lamp, loudspeaker, rifle, sofa, table, telephone, and vessel}), resulting in a total of 13,000 instances. We build two separate test sets for evaluation. The first test set is the ``novel instances'' set, which contains \nguyen{50 new instances for each training category}. The second test set is the ``novel category'' set, which includes \nguyen{100 models per category for} the 10 unseen categories selected by FORGE~(\textit{bus, guitar, clock, bottle, train, mug, washer, skateboard, dishwasher, and pistol}). For each 3D model, we randomly select camera poses to produce five reference images and five query images. We use BlenderProc~\cite{denninger2019blenderproc} as rendering engine. 

Figure~\ref{fig:shapeNet} illustrates the categories used for training our architecture and the categories used for testing it. The shapes and appearances of the categories in the test set are very different from the shapes and appearances of the categories in the training set, and thus constitute a good test set for generalization to unseen categories.

\vspace{-12pt}
\paragraph{Real-world dataset.}
We evaluate on the T-LESS dataset \cite{hodan-wacv17-tless} following the evaluation protocol of \cite{sundermeyer-cvpr20-multipathlearning}: we train only on objects 1-18 and test on the full PrimeSense test set using the ground-truth masks. At inference, we randomly sample a non-occluded reference image either from \emph{all views} or only from \emph{front views}~(-45\textdegree $\leq$ azimuth $\leq$ 45\textdegree), which often offers more information on the object and illustrates the influence of the reference view.


\vspace{-12pt}
\paragraph{Metrics.}
For the ShapeNet dataset, we report two different metrics based on relative camera pose error as done in \cite{mariotti_semi-supervised_2020}. Specifically, we provide the median pose error across instances for each category in the test set, and the accuracy $\bm{\AccThirty}$ for which a prediction is treated as correct when the pose error is $\le 30^{\circ}$. Additionally, we present the results of our method for the top 3 and 5 nearest neighbors retrieved by template matching.

For the T-LESS dataset, as most objects are symmetric, we report the recall VSD metric as done in \cite{sundermeyer-cvpr20-multipathlearning}. Please note that for the evaluation on the T-LESS dataset, we also predict the translation by using the same formula ``projective distance estimation'' as SSD-6D~\cite{kehl-iccv17-ssd6d}, as done in~\cite{sundermeyer-eccv18-implicit3dorientationlearning, sundermeyer-cvpr20-multipathlearning}. This translation is deduced from the retrieved template and the relative scale factor between the two input images, as detailed in Section~8 of~\cite{nguyen2022templates}.


\vspace{-12pt}
\paragraph{Baselines.}

We compare our work with all previous methods that aim to predict a pose from a single view: PIZZA \cite{nguyen_pizza_2022}, a regression-based approach that directly predicts the relative pose, as well as SSVE~\cite{mariotti_semi-supervised_2020} and ViewNet~\cite{mariotti_viewnet_2021}, which employ semi-supervised and self-supervised techniques to treat viewpoint estimation as an image reconstruction problem using conditional generation. We also compare our method with the recent diffusion-based method 3DiM~\cite{3dim}, which generates pixel-level view synthesis. Since 3DiM originally only targets view-synthesis and is not designed for 3D object pose, we use it to generate templates and perform nearest neighbor search to estimate a 3D object pose. To make 3DiM work in the same setting as us, we retrain it using relative pose conditioning instead of canonical pose conditioning.

\vspace{-12pt}
\paragraph{Implementation.}
Only the code of PIZZA is available. The other methods did not release their code at the time of writing, however we re-implemented them. We use a ResNet18 backbone  as in~\cite{nguyen_pizza_2022} for PIZZA, SSVE, and ViewNet. We train all models on input images with a resolution of 256$\times$256 except for 3DiM for which we use a resolution of 128$\times$128 since 3DiM performs view synthesis in pixel space, which takes much more memory. Our re-implementations achieve similar performance as the original papers when evaluated on the same data for seen categories, as shown in Table~\ref{tab:shapeNet}, which validates our comparisons. 
Our method also uses the frozen encoder from \cite{stable-diffusion} to encode the input images into embeddings of size 32$\times$32$\times$8. \nguyen{In all settings, we train the baselines and our method using the same training set and  AdamW~\cite{loshchilov_decoupled_2017} with an initial learning rate of $5\,{\times}\,10^{-5}$. 
Training takes about 20 hours on 4 V100 GPUs for each method.

} 

\begin{table}[t]
\centering
\resizebox{\linewidth}{!}{
\begin{tabular}{@{\,}l | l l c c c@{\,}}
\toprule
 & \multirow{2}{*}{\raisebox{-1mm}{\bf Method}} & 
 \multirow{2}{*}{\begin{tabular}{@{}c@{}} \textbf{Ref.\ image}\raisebox{4mm}{} \\ \textbf{sampling} \end{tabular}}
 & \multicolumn{3}{c}{\textbf{Recall VSD}} \\
 \cmidrule(l){4-6}
& & 
& \textbf{Seen obj.} 
& \!\!\!\!\textbf{Novel obj.}\!\!\!\! 
& \textbf{Avg} \\
 \midrule
  \multirow{2}{*}{\rotatebox[origin=c]{90}{\stackbox[c][b]{\baselineskip=12pt GT\\CAD\par}}}

  & Nguyen et al.~\cite{nguyen2022templates} & -  & \bf 60.15 & \bf 58.70 & \bf 59.57\\
  & MultiPath~\cite{sundermeyer-cvpr20-multipathlearning} & - & 43.17 & 43.33 & 43.24\\
  \midrule
  \multirow{2}{*}{\rotatebox[origin=c]{90}{\stackbox[c][b]{\baselineskip=12pt 1 ref.~image~~\\(avg 5 runs)~~\par}}}
  & PIZZA~\cite{nguyen_pizza_2022} & all views & 20.05 & 15.90 & 18.39  \\
  & Ours & all views & \bf 47.03 & \bf 45.69 & \bf 46.49 \\
 \cmidrule(lr){2-6}
  & PIZZA~\cite{nguyen_pizza_2022} & front views & 21.63 & 15.55 & 19.19 \\
  & Ours & front views & \bf 49.30 & \bf 48.46 & \bf 48.96 \\
\bottomrule
\end{tabular}}
\caption{{\nguyen{\bf Comparison to PIZZA~\cite{nguyen_pizza_2022} and CAD-based methods~\cite{nguyen2022templates, sundermeyer-cvpr20-multipathlearning}} on seen (obj.~1-18) and novel (obj.~19-30) objects of T-LESS.
We report  numbers averaged over 5 different samplings and runs.}}
\label{tab:tless}
\end{table}

\begin{figure}[t]
    \centering
    \begin{tabular}{cc}
    \footnotesize{Seen objects: \#4, \#14} & \footnotesize{Novel objects: \#20, \#22} \\
    \includegraphics[width=0.46\linewidth]{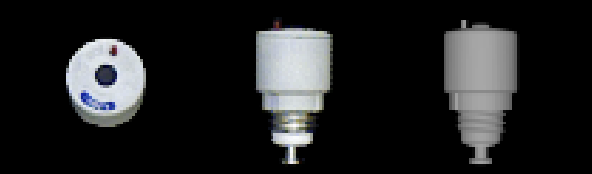} & \includegraphics[width=0.46\linewidth]{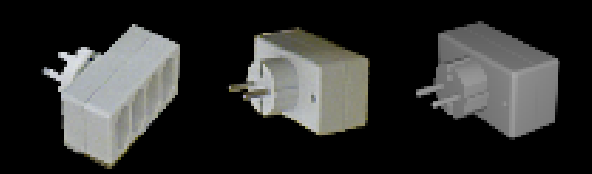}\\
    \includegraphics[width=0.46\linewidth]{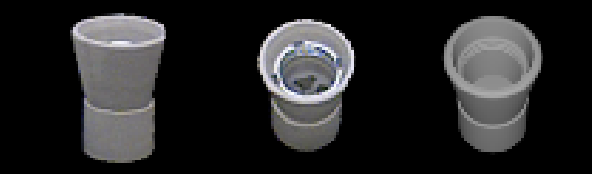} & \includegraphics[width=0.46\linewidth]{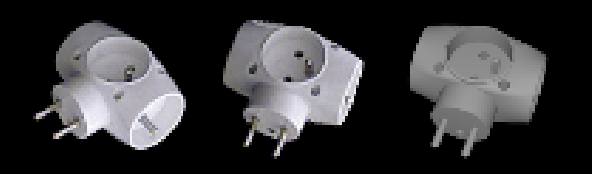} \\[-1mm]
    \footnotesize{Reference$\;\;\;\;$Query$\;\;\;\;$Prediction} & \footnotesize{Reference$\;\;\;\;$Query$\;\;\;\;$Prediction}\\
    \end{tabular}
    \caption{{\bf Qualitative results on real images of T-LESS.} For each sample, we show in the last column the predicted poses. }
    \label{fig:tless}
\end{figure}

\begin{figure}[t]
\setlength\plotHeight{2.0cm}
\centering
\setlength\lineskip{1.5pt}
\setlength\tabcolsep{1.5pt} 
{\footnotesize
\begin{tabular}{cr}
\begin{tabular}{
>{\centering\arraybackslash}m{\plotHeight}
>{\centering\arraybackslash}m{\plotHeight}
>{\centering\arraybackslash}m{\plotHeight}
>{\centering\arraybackslash}m{\plotHeight}
}
\frame{\includegraphics[height=\plotHeight, ]{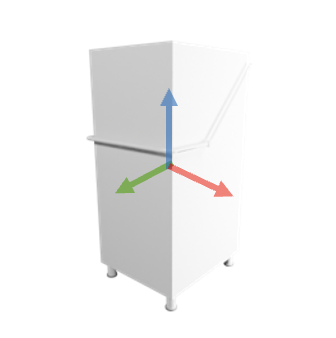}}&
\frame{\includegraphics[height=\plotHeight, ]{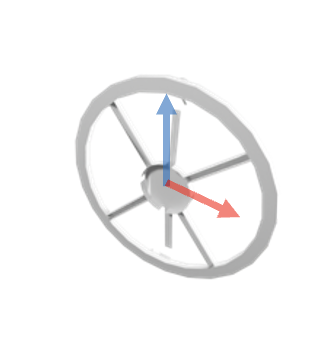}}&

\frame{\includegraphics[height=\plotHeight, ]{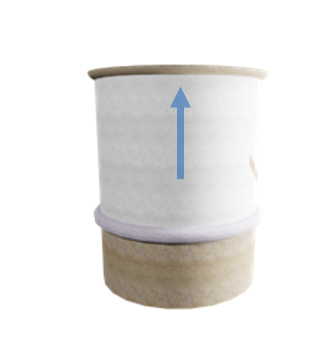}}&
\frame{\includegraphics[height=\plotHeight, ]{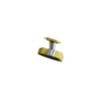}}\\
\footnotesize{\;\;\;\;\;Dishwasher$\;\;\;\;\;\;\;\;\;\;\;\;\;$Clock$\;\;\;\;\;\;\;\;\;\;\;\;\;\;\;\;\;\;\;$Mug\;\;\;\;\;\;\;\;\;\;\;\;\;\;\;\;\;\;\;Guitar}
\end{tabular}
\end{tabular}}
\caption{\nguyen{{\bf Failure cases.} ``Dishwashers'', ``clocks'', and ``dishwashers'' are ``nearly symmetrical'' while ``guitars'' are barely visible from some viewpoints. This makes the pose estimation very challenging, and all the methods perform poorly on these categories.}}
\label{fig:failureCases}
\end{figure}

\begin{figure*}
\newlength{\imageheight}
\setlength\imageheight{1.48cm}
\centering
\setlength\lineskip{1.pt}
\setlength\tabcolsep{1.pt} 
{\small
\begin{tabular}{c}
\begin{tabular}{
>{\centering\arraybackslash}m{0.5cm}
>{\centering\arraybackslash}m{\imageheight}
>{\centering\arraybackslash}m{\imageheight}
>{\centering\arraybackslash}m{\imageheight}
>{\centering\arraybackslash}m{\imageheight}
>{\centering\arraybackslash}m{\imageheight}
>{\centering\arraybackslash}m{1.cm}
>{\centering\arraybackslash}m{\imageheight}
>{\centering\arraybackslash}m{\imageheight}
>{\centering\arraybackslash}m{\imageheight}
>{\centering\arraybackslash}m{\imageheight}
>{\centering\arraybackslash}m{\imageheight}
}
\hline
\multicolumn{12}{c}{\textbf{without occlusions}}\\
\hline
\\[-0.2cm]
\parbox[t]{0mm}{\rotatebox[origin=c]{90}{Bottle}} & \frame{\includegraphics[height=\imageheight, ]{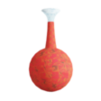}} &
\frame{\includegraphics[height=\imageheight, ]{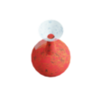}} &
\frame{\includegraphics[height=\imageheight, ]{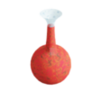}}&
\frame{\includegraphics[height=\imageheight, ]{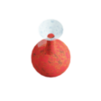}}&
\frame{\includegraphics[height=\imageheight, ]{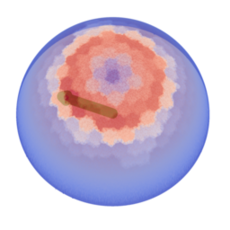}}&
&
\frame{\includegraphics[height=\imageheight, ]{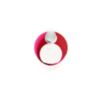}} &
\frame{\includegraphics[height=\imageheight, ]{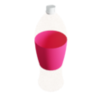}} &
\frame{\includegraphics[height=\imageheight, ]{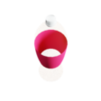}}&
\frame{\includegraphics[height=\imageheight, ]{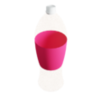}}&
\frame{\includegraphics[height=\imageheight, ]{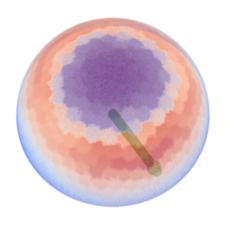}}\\


\parbox[t]{0mm}{\rotatebox[origin=c]{90}{Clock}} & \frame{\includegraphics[height=\imageheight, ]{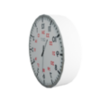}} &
\frame{\includegraphics[height=\imageheight, ]{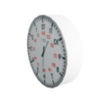}} &
\frame{\includegraphics[height=\imageheight, ]{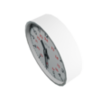}}&
\frame{\includegraphics[height=\imageheight, ]{figures/experiments/qualitative/clock_138_31_9/0_pred_crop.png}}&
\frame{\includegraphics[height=\imageheight, ]{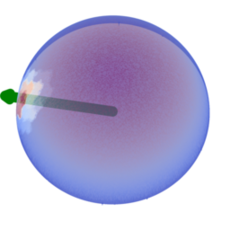}}&
&
\frame{\includegraphics[height=\imageheight, ]{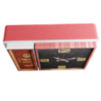}} &
\frame{\includegraphics[height=\imageheight, ]{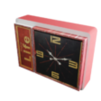}} &
\frame{\includegraphics[height=\imageheight, ]{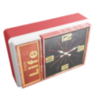}}&
\frame{\includegraphics[height=\imageheight, ]{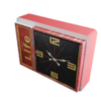}}&
\frame{\includegraphics[height=\imageheight, ]{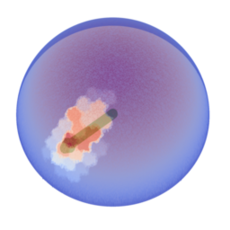}}\\

\parbox[t]{0mm}{\rotatebox[origin=l]{90}{Dishwasher}} & \frame{\includegraphics[height=\imageheight, ]{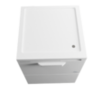}} &
\frame{\includegraphics[height=\imageheight, ]{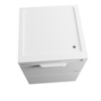}} &
\frame{\includegraphics[height=\imageheight, ]{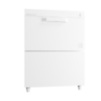}}&
\frame{\includegraphics[height=\imageheight, ]{figures/experiments/qualitative/dishwasher_167_3_8/0_pred_crop.png}}&
\frame{\includegraphics[height=\imageheight, ]{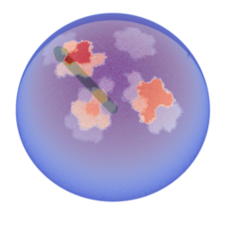}}&
&
\frame{\includegraphics[height=\imageheight, ]{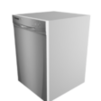}} &
\frame{\includegraphics[height=\imageheight, ]{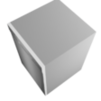}} &
\frame{\includegraphics[height=\imageheight, ]{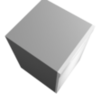}}&
\frame{\includegraphics[height=\imageheight, ]{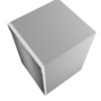}}&
\frame{\includegraphics[height=\imageheight, ]{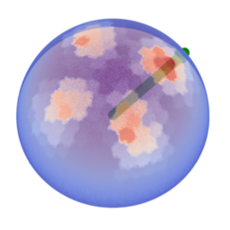}}\\

\parbox[t]{0mm}{\rotatebox[origin=c]{90}{Guitar}} & \frame{\includegraphics[height=\imageheight, ]{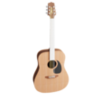}} &
\frame{\includegraphics[height=\imageheight, ]{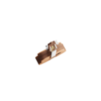}} &
\frame{\includegraphics[height=\imageheight, ]{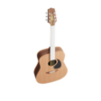}}&
\frame{\includegraphics[height=\imageheight, ]{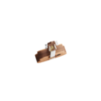}}&
\frame{\includegraphics[height=\imageheight, ]{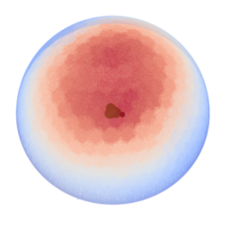}}&
&
\frame{\includegraphics[height=\imageheight, ]{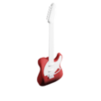}} &
\frame{\includegraphics[height=\imageheight, ]{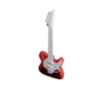}} &
\frame{\includegraphics[height=\imageheight, ]{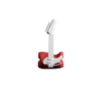}}&
\frame{\includegraphics[height=\imageheight, ]{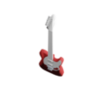}}&
\frame{\includegraphics[height=\imageheight, ]{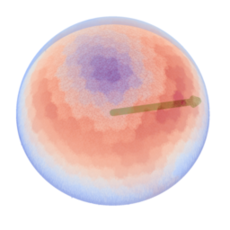}}\\

\parbox[t]{0mm}{\rotatebox[origin=c]{90}{Mug}} & \frame{\includegraphics[height=\imageheight, ]{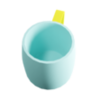}} &
\frame{\includegraphics[height=\imageheight, ]{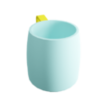}} &
\frame{\includegraphics[height=\imageheight, ]{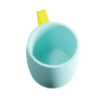}}&
\frame{\includegraphics[height=\imageheight, ]{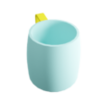}}&
\frame{\includegraphics[height=\imageheight, ]{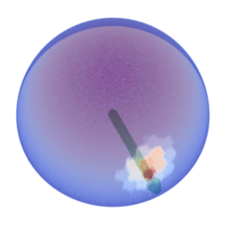}}&
&
\frame{\includegraphics[height=\imageheight, ]{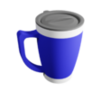}} &
\frame{\includegraphics[height=\imageheight, ]{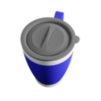}} &
\frame{\includegraphics[height=\imageheight, ]{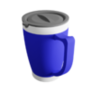}}&
\frame{\includegraphics[height=\imageheight, ]{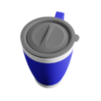}}&
\frame{\includegraphics[height=\imageheight, ]{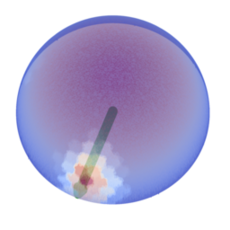}}\\

\parbox[t]{0mm}{\rotatebox[origin=c]{90}{Pistol}} & \frame{\includegraphics[height=\imageheight, ]{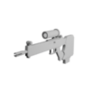}} &
\frame{\includegraphics[height=\imageheight, ]{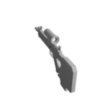}} &
\frame{\includegraphics[height=\imageheight, ]{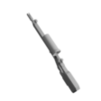}}&
\frame{\includegraphics[height=\imageheight, ]{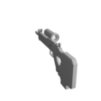}}&
\frame{\includegraphics[height=\imageheight, ]{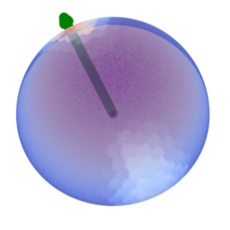}}&
&
\frame{\includegraphics[height=\imageheight, ]{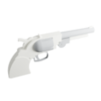}} &
\frame{\includegraphics[height=\imageheight, ]{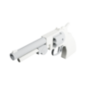}} &
\frame{\includegraphics[height=\imageheight, ]{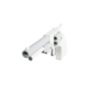}}&
\frame{\includegraphics[height=\imageheight, ]{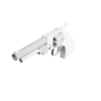}}&
\frame{\includegraphics[height=\imageheight, ]{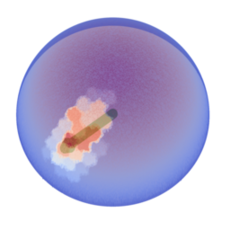}}\\

\parbox[t]{1mm}{\rotatebox[origin=l]{90}{Skateboard}} & \frame{\includegraphics[height=\imageheight, ]{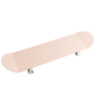}} &
\frame{\includegraphics[height=\imageheight, ]{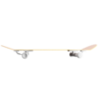}} &
\frame{\includegraphics[height=\imageheight, ]{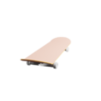}}&
\frame{\includegraphics[height=\imageheight, ]{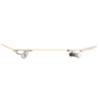}}&
\frame{\includegraphics[height=\imageheight, ]{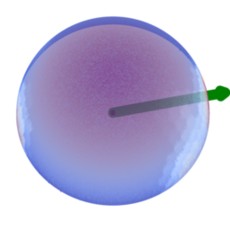}}&
&
\frame{\includegraphics[height=\imageheight, ]{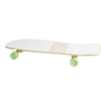}} &
\frame{\includegraphics[height=\imageheight, ]{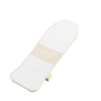}} &
\frame{\includegraphics[height=\imageheight, ]{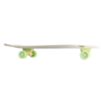}}&
\frame{\includegraphics[height=\imageheight, ]{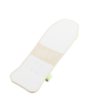}}&
\frame{\includegraphics[height=\imageheight, ]{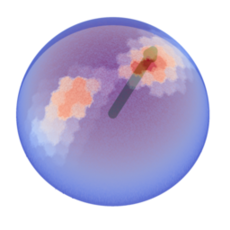}}\\

\parbox[t]{1mm}{\rotatebox[origin=c]{90}{Train}} & \frame{\includegraphics[height=\imageheight, ]{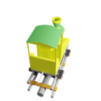}} &
\frame{\includegraphics[height=\imageheight, ]{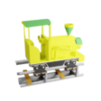}} &
\frame{\includegraphics[height=\imageheight, ]{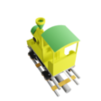}}&
\frame{\includegraphics[height=\imageheight, ]{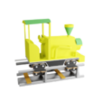}}&
\frame{\includegraphics[height=\imageheight, ]{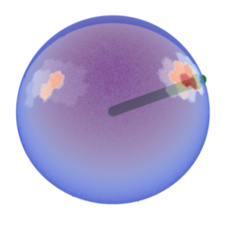}}&
&
\frame{\includegraphics[height=\imageheight, ]{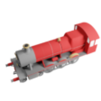}} &
\frame{\includegraphics[height=\imageheight, ]{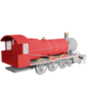}} &
\frame{\includegraphics[height=\imageheight, ]{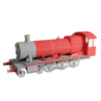}}&
\frame{\includegraphics[height=\imageheight, ]{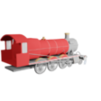}}&
\frame{\includegraphics[height=\imageheight, ]{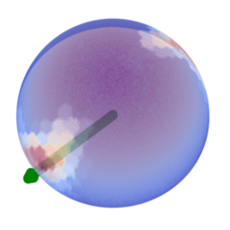}}\\[0.1cm]

\parbox[t]{1mm}{\rotatebox[origin=c]{90}{Washer}} & \frame{\includegraphics[height=\imageheight, ]{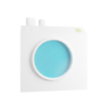}} &
\frame{\includegraphics[height=\imageheight, ]{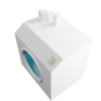}} &
\frame{\includegraphics[height=\imageheight, ]{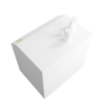}}&
\frame{\includegraphics[height=\imageheight, ]{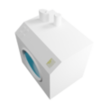}}&
\frame{\includegraphics[height=\imageheight, ]{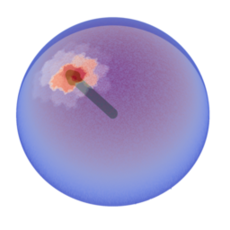}}&
&
\frame{\includegraphics[height=\imageheight, ]{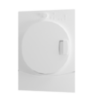}} &
\frame{\includegraphics[height=\imageheight, ]{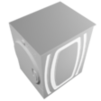}} &
\frame{\includegraphics[height=\imageheight, ]{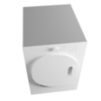}}&
\frame{\includegraphics[height=\imageheight, ]{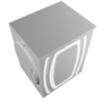}}&
\frame{\includegraphics[height=\imageheight, ]{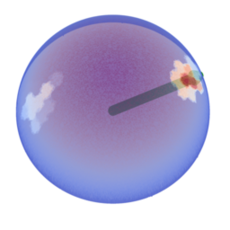}}\\

\hline
\multicolumn{12}{c}{\textbf{with partial occlusions}}\\
\hline
\\[-0.2cm]

& \frame{\includegraphics[height=\imageheight, ]{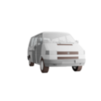}} &
\frame{\includegraphics[height=\imageheight, ]{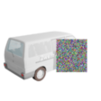}} &
\frame{\includegraphics[height=\imageheight, ]{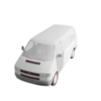}}&
\frame{\includegraphics[height=\imageheight, ]{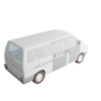}}&
\frame{\includegraphics[height=\imageheight, ]{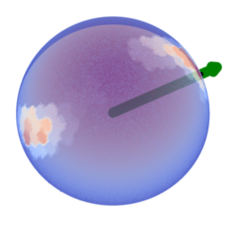}}&
&

\frame{\includegraphics[height=\imageheight, ]{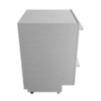}} &
\frame{\includegraphics[height=\imageheight, ]{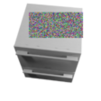}} &
\frame{\includegraphics[height=\imageheight, ]{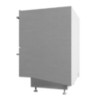}}&
\frame{\includegraphics[height=\imageheight, ]{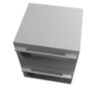}}&
\frame{\includegraphics[height=\imageheight, ]{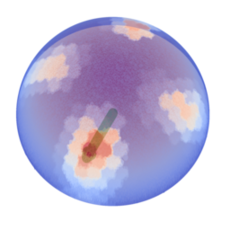}}\\

& \frame{\includegraphics[height=\imageheight, ]{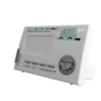}} &
\frame{\includegraphics[height=\imageheight, ]{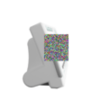}} &
\frame{\includegraphics[height=\imageheight, ]{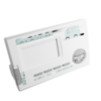}}&
\frame{\includegraphics[height=\imageheight, ]{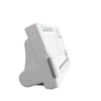}}&
\frame{\includegraphics[height=\imageheight, ]{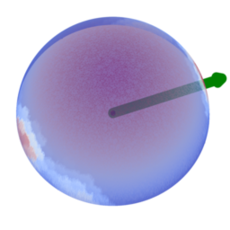}}&
&

\frame{\includegraphics[height=\imageheight, ]{figures/experiments/qualitative_occlusion/clock_142_35_7/ref.png}} &
\frame{\includegraphics[height=\imageheight, ]{figures/experiments/qualitative_occlusion/clock_142_35_7/query.png}} &
\frame{\includegraphics[height=\imageheight, ]{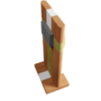}}&
\frame{\includegraphics[height=\imageheight, ]{figures/experiments/qualitative_occlusion/clock_142_35_7/0_pred_crop.png}}&
\frame{\includegraphics[height=\imageheight, ]{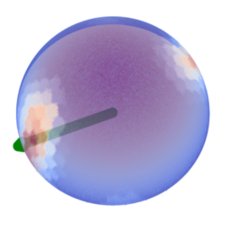}}\\

& \frame{\includegraphics[height=\imageheight, ]{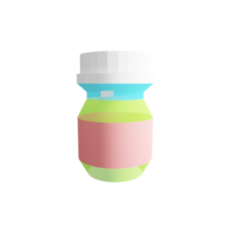}} &
\frame{\includegraphics[height=\imageheight, ]{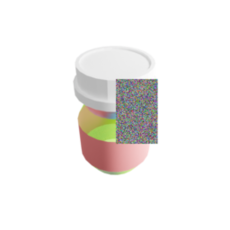}}&
\frame{\includegraphics[height=\imageheight, ]{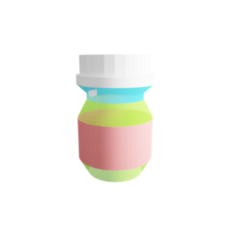}}&
\frame{\includegraphics[height=\imageheight, ]{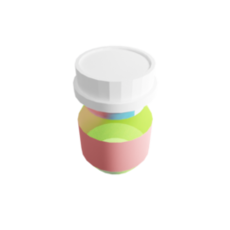}}&
\frame{\includegraphics[height=\imageheight, ]{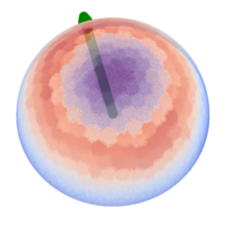}}&

&
\frame{\includegraphics[height=\imageheight, ]{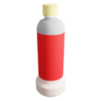}} &
\frame{\includegraphics[height=\imageheight, ]{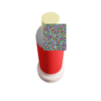}} &
\frame{\includegraphics[height=\imageheight, ]{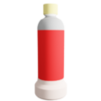}}&
\frame{\includegraphics[height=\imageheight, ]{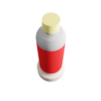}}&
\frame{\includegraphics[height=\imageheight, ]{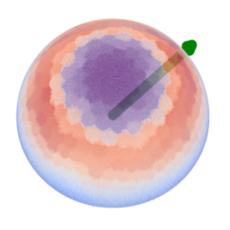}}\\

&Reference  & Query & PIZZA~\cite{nguyen_pizza_2022} & Ours & Pose distribution & &Reference  & Query & PIZZA~\cite{nguyen_pizza_2022} & Ours & Pose distribution\\
\end{tabular}
\end{tabular}
}
\vspace{-3mm}
\caption{\label{fig:qualitative} {\bf Visual results on unseen categories} from ShapeNet. An arrow indicates the pose with the highest probability as recovered by our method. We visually compare with PIZZA, which is the method with the second best performance. \textbf{We visualize the predicted poses by rendering the object from these poses, but the 3D model is only used for visualization purposes, not as input to our method. Similarly, we use the canonical pose of the 3D model to visualize this distribution, but not as input to our method.}
} 
\end{figure*}

\subsection{Comparison with the state of the art}
\label{sec:main_results}
\subsubsection{Results on ShapeNet}
Table~\ref{tab:shapeNet} summarizes the results of our method compared with the baselines discussed above.
Under both the Acc30 and Median metrics, our method consistently achieves the best overall performance, outperforming the baselines by more than 10\% in Acc30 and 10$^o$ in Median. 
In particular, while other works produce reasonable results on unseen instances of seen training categories, they often struggle to estimate the 3D pose of objects from unseen categories. By contrast, our method works well in this case, demonstrating a better generalization ability on unseen categories.


Figure~\ref{fig:qualitative} shows some visualization results of our method on unseen categories, with and without symmetries. Our method  produces more accurate 3D poses than the baselines when there is a symmetry axis.

\subsubsection{Results on T-LESS}

Table~\ref{tab:tless} shows our comparison with~\cite{nguyen_pizza_2022,sundermeyer-cvpr20-multipathlearning,nguyen2022templates} on real images of T-LESS. While our method focuses on the more challenging case of using \textit{a single reference image}, \cite{nguyen2022templates,sundermeyer-cvpr20-multipathlearning} rely on ground-truth CAD models. Our method consistently outperforms the baseline PIZZA by a large margin. Interestingly, although there is still a gap compared to the SOTA~\cite{nguyen2022templates}, our method outperforms MultiPath~\cite{sundermeyer-cvpr20-multipathlearning}. Figure~\ref{fig:tless} shows results on seen and unseen objects of T-LESS.

\subsection{Robustness to occlusions}
\label{sec:occlusions}
To evaluate the robustness of our method against occlusions, we added random rectangle filled with Gaussian noise to the query images over the objects, in a  similar way to  Random Erasing~\cite{zhong2020random}. We vary the size of the rectangles to cover a range betwen 0\% to 25\% of the bounding box of the object.  Figures~\ref{fig:teaser} and \ref{fig:qualitative} show several examples.

\begin{table}[!t]
    \addtolength{\tabcolsep}{-2pt}
    \centering
    \scalebox{0.85}{
    \begin{tabular}{@{}l l c c  c  c  c  c }
    \toprule
    \parbox[t]{2mm}{\multirow{2}{*}{\rotatebox[origin=c]{90}{$\bm{\AccThirty}\uparrow$}}} & Method &  0\% & 5\% & 10\% & 15\%  & 20\% & 25\% \\
    \cmidrule{2-8}
    & PIZZA \cite{nguyen_pizza_2022} & 48.9 &  44.6 &  33.3 &  24.5 &  18.2 &  14.6 \\
    & NOPE (ours) & \bf 59.8 & \bf 54.3 &  \bf 48.4 &  \bf 45.1 &  \bf 43.7 &  \bf 40.5  \\
    \bottomrule
    \end{tabular}}
    \caption{{\bf Robustness to partial occlusions. } We add rectangles of Gaussian noise to the query image, and vary the ratio between the area of the rectangle and the area of the object's 2D bounding box. 
    Our method remains robust under large occlusions, while PIZZA's performance decreases significantly.
    }
    \label{tab:occlusion}
\end{table}

Table~\ref{tab:occlusion} compares PIZZA, the best second performing method in our previous evaluation, to our method for different occlusion rates. Our method remains robust even under large occlusions, thanks to embedding matching. Figure~\ref{fig:qualitative} shows that our pose probabilities remain peaked on the correct maximum and shows clearly the symmetries.

\begin{table}[t]
	\centering
	\resizebox{0.88\linewidth}{!}{
	\begin{tabular}{@{}l r c c@{}}
	\toprule
    \multirow{2}{*}{\bf Method} 
    &\multirow{2}{*}{\bf \parbox{1.3cm}{Memory}}
     & 
    \multicolumn{2}{c}{\textbf{Run-time}}\\
	\cmidrule(lr){3-4}                                
	& & \textbf{Processing} & \textbf{Neighbors search} \\
	\midrule
	3DiM \cite{3dim} & 358.6 MB &  13 min &   0.31 s\\
	NOPE (ours) &  22.4 MB &  1.01 s &  0.18 s\\
    \bottomrule
	\end{tabular}}
\caption{{\bf Average run-time}  of our method and 3DiM \cite{3dim} on a single GPU V100. We report the memory used for storing novel views, the time taken to generate novel views, and the time taken for nearest neighbor search to obtain the final prediction. }
    \label{tab:runtime}
\end{table}

\subsection{Runtime analysis} 
\label{sec:run_time}
We report the running time of NOPE and 3DiM in Table~\ref{tab:runtime}. Our method is significantly faster than 3DiM, thanks to our strategy of predicting the embedding of novel viewpoints with a single step instead of multiple diffusion steps.

\subsection{Failure cases}
\label{sec:failureCases}


All the methods fail to yield accurate results when evaluated on ``clock'', ``dishwasher'', ``guitar'', and ``mug'' categories, as indicated by the high median errors. As shown in Figure~\ref{fig:failureCases}, these categories except ``guitar'' are ``almost symmetric'', in the sense that only small details make the pose non-ambiguous. Our predictions using the top-3 and top-5 nearest neighbors significantly improves median errors for 90-symmetrical, 180-symmetrical objects, but not circular-symmetrical as mug objects. Additionally, guitar objects  can appear very thin under certain viewpoints.

\section{Conclusion}


\label{sec:conclusion}

Our experiments have shown that direct inference of average view embeddings from a single view, as in NOPE, leads to accurate object pose estimation. This is true even for objects from unseen categories, while requiring neither retraining nor a 3D model. NOPE also lets us estimate the pose ambiguities that arise for many objects.

}

{\small \noindent\textbf{Acknowledgments.} The authors thank Elliot Vincent, Mathis Petrovich and Nicolas Dufour for helpful discussions. This project was funded in part by the European Union (ERC Advanced Grant explorer  Funding ID \#101097259). This work was performed using HPC resources from GENCI–IDRIS 2022-AD011012294R2 and 2022-AD011012294R3. 

{\small
\bibliographystyle{ieee_fullname}
\bibliography{cleaned_refs}
}

\end{document}